\documentclass[fleqn,10pt]{wlscirep}
\usepackage[utf8]{inputenc}
\usepackage[T1]{fontenc}
\usepackage{xcolor}
\usepackage{subcaption}

\title{Scalable neuromorphic computing from autonomous spiking dynamics in a clockless reconfigurable chip}

\author[1,*]{Eric Oliveira Gomes}
\author[1,*]{Damien Rontani}
\affil[1]{LMOPS UR4423 Laboratory, CentraleSup\'elec and Universit\'e de Lorraine, Metz F-57070, France}
\affil[*]{eric.oliveira-gomes@centralesupelec.fr - damien.rontani@centralesupelec.fr}


\begin{abstract}
We propose a scalable neuromorphic architecture based on spiking dynamics emerging from the autonomous time-continuous evolution of clockless (asynchronous) digital circuits. Implemented on commercially available field-programmable gate arrays (FPGAs), our system implements networks of interacting Boolean spiking neurons with configurable excitatory and inhibitory synaptic weights. A complete processing pipeline enables efficient handling of spike-encoded data for solving machine-learning tasks. We demonstrate competitive performance for an audio classification task with spike-based encoding and high-speed processing. Power consumption is significantly lower than traditional digital implementations;  this makes our approach an efficient alternative that bridges the gap to dedicated analog neuromorphic systems without the need for specialized hardware design. More generally, our approach establishes clockless digital hardware as a viable platform for neuromorphic computing. It paves the way for reconfigurable chips to be turned into energy-efficient quasi-analog neuromorphic processors.
\end{abstract}

\begin{document}
\flushbottom
\maketitle
\thispagestyle{empty}

\section*{Introduction}
Neuromorphic technologies are known for their capacity to emulate neural dynamics found in biological neural networks by exploiting specific properties of physical devices. Amongst the chief advantages of these approaches, there are their large bandwidth, scalability and high energy efficiency, outperforming their digital software counterparts by several orders of magnitude \cite{Ostrau_benchmarking_2022,Pal_ultraenergyeff_2024}. As a result, they constitute a promising alternative for the backbone of future infrastructure that will support an ever increasing demand for artificial-intelligence (AI) solutions.

Over the last decades, there have been intensive research and proposals to use various physical media to embody artificial neural networks (ANN) especially within the paradigm of reservoir computing (RC)\cite{Jaeger2004,Maass2002}, which allows for a simplified training framework of recurrent neural networks (RNN) \cite{Lukosevicius2009}. Physical ANN relying on photonics\cite{Lugnan2020}, electronics \cite{Appeltant2011,Du2017}, and spintronics\cite{Torrejon2017,Romera2018} have been successfully demonstrated and solved machine learning tasks such as multimedia signal classification \cite{Antonik_ImageClassif_2020,Milano_speech_2022} or regression-like tasks such as time-series prediction \cite{Shahi_prediction_2022}. These implementations, however, do not necessarily focus on the biological relevance of their dynamics, but rather on the possibility to nonlinearly process high-dimensional input data at high speed, with consistency and sufficient signal-to-noise ratio.

Spiking neural networks (SNN) are a specific subset of ANN, where the artificial neurons dynamics should closely resemble that of biological neurons \cite{Yamazaki2022}. They can be challenging to implement because triggering the emission of physical pulses and obtaining sufficiently rich dynamics requires specific dynamical features such as excitability \cite{excitability_Izhikevich2000}, the existence of a refractory mechanism \cite{moosavi_refractory_2017}, inhibition \cite{IsaacsonInhibition2011}, or more advanced features such as spike-timing-dependent plasticity (STDP)\cite{song_STDP_2000}. This will either require the algorithmic simulation of neuron models \cite{Gerstner_book_2014} via digital logic on conventional hardware or carefully engineering novel devices with innovative materials to display such biological features \cite{Sangwan_neuromorphic_2020, dagostinoDenRAMNeuromorphicDendritic2024}. Currently, there is a combined effort in both the development of hardware solutions and algorithms to enable spike-based computing \cite{roy_towards_2019}.

A wide variety of methods have been explored to construct systems capable of efficiently emulating SNNs. Among them, large-scale architectures such as Intel Loihi \cite{davies_loihi_2018}, IBM TrueNorth \cite{akopyan_truenorth_2015}, SpiNNaker \cite{furber_spinnaker_2014}, and Tianjic \cite{deng_tianjic_2020} employ digital electronic designs optimized for the simulation of spiking neural networks. Likewise, large-scale mixed-signal electronic designs have also been proposed, such as Neurogrid \cite{benjamin_neurogrid_2014} and BrainScaleS-2 \cite{pehle_brainscales-2_2022}, which avoid the computational cost of evaluating neuron models by instead using analog circuits that directly reproduce their dynamical behavior. In addition to electronic implementations, a wide variety of physical phenomena have been explored to create neuromorphic systems \cite{markovic_physics_2020}, including photonic approaches \cite{shastri_photonics_2021,prucnal_recent_2016,talukder_spiking_2025} and resistive devices such as memristors \cite{indiveri_integration_2013,wang_fully_2018}. Nevertheless, a limitation of these architectures is the requirement for specialized hardware, which has motivated the development of neuron implementations on reconfigurable electronic platforms such as field-programmable gate arrays (FPGA) and field-programmable analog arrays (FPAA).

FPGA devices are commercially available reconfigurable microelectronic chips traditionally used for digital circuit validation and signal-processing tasks. The ample flexibility of these devices has been leveraged to create a wide variety of digital architectures capable of simulating SNNs more efficiently than a conventional central processing unit (CPU) \cite{farsa_reconfigurable_2025, peng_li_fpga_implementation}. Similarly, FPAA platforms, which share several features with FPGAs but include analog components that enable mixed-signal processing \cite{hasler_large-scale_2020}, have also been successfully used to construct neuron circuits \cite{natarajan_hodgkinhuxley_2018,bhattacharyya_six-transistor_2025}. Additionally, the analog behavior of FPGA devices has been exploited for machine learning tasks \cite{haynes_reservoir_2015, shani_dynamics_2019}, although not with a spike-based neuromorphic system.

In their seminal work, Rosin et al. proposed an elementary excitable system exploiting autonomous Boolean network (ABN) dynamics that emerge from removing the master clock in electronic logical circuits \cite{Rosin2012}. Clockless digital circuits become time-continuous and can show dynamics not accessible to discrete-time finite-state machines (\emph{e.g.} deterministic chaos \cite{Zhang2009,Rosin2013}). For neuromorphic architectures, and contrary to the aforementioned approaches requiring the development of a dedicated hardware solution, it was shown that autonomous dynamics could be harnessed to generate nanosecond pulsing dynamics by unclocking inexpensive FPGA devices. These excitable systems were used as a framework to observe various collective phenomena in purely excitatory complex networks, such as cluster synchronization \cite{Rosin2013b}. However, they lacked key neuromorphic capabilities, such as the integration of presynaptic spikes and the ability to implement inhibitory behavior, which is a key property to ensure balanced spiking dynamics in neural networks \cite{IsaacsonInhibition2011}.

In this work, we introduce a novel autonomous Boolean neuronal architecture, allowing for spike integration, as well as excitatory and inhibitory signals. We also present methods for implementing network-embedded synaptic weights and delays. This architecture enables the deployment of densely interconnected SNNs on reconfigurable hardware, such as FPGAs, while operating in a high-speed, quasi-analog regime and exploiting the physical response of FPGA's transistors. We also propose an advanced digital front-end, implemented in the FPGA chip, enabling controlled network stimulation and the subsequent acquisition of spiking activity. This novel neuromorphic approach can be easily deployed on low-cost reconfigurable microchips with large processing bandwidth. Furthermore, we demonstrate how to leverage our architecture as a liquid state machine (LSM) \cite{Maass2002} for machine-learning applications. The efficacy of this approach is validated through the classification of audio signals from the Spiking Heidelberg Digits (SHD) dataset~\cite{cramerHeidelbergSpikingData2022}. Our approach shows great potential toward accelerating the development of quasi-analog neuromorphic architectures without resorting to proprietary dedicated hardware solutions.

\section*{Results}
In this section, we first describe the structure of our spiking Boolean neuron built with autonomous (clockless) digital logic elements from an FPGA reconfigurable chip. Then, we detail how to create biologically-relevant spiking neural networks balancing excitation and inhibition, and also incorporating propagation delays. Finally, we detail the implementation of an integrated machine-learning pipeline to exploit our autonomous Boolean SNNs as LSMs to solve classification tasks with spike-encoded input data.  

\subsubsection*{Structure of a Spiking Boolean Neuron}
\begin{figure}[t!]
\centering
\includegraphics[width=1.0\linewidth]{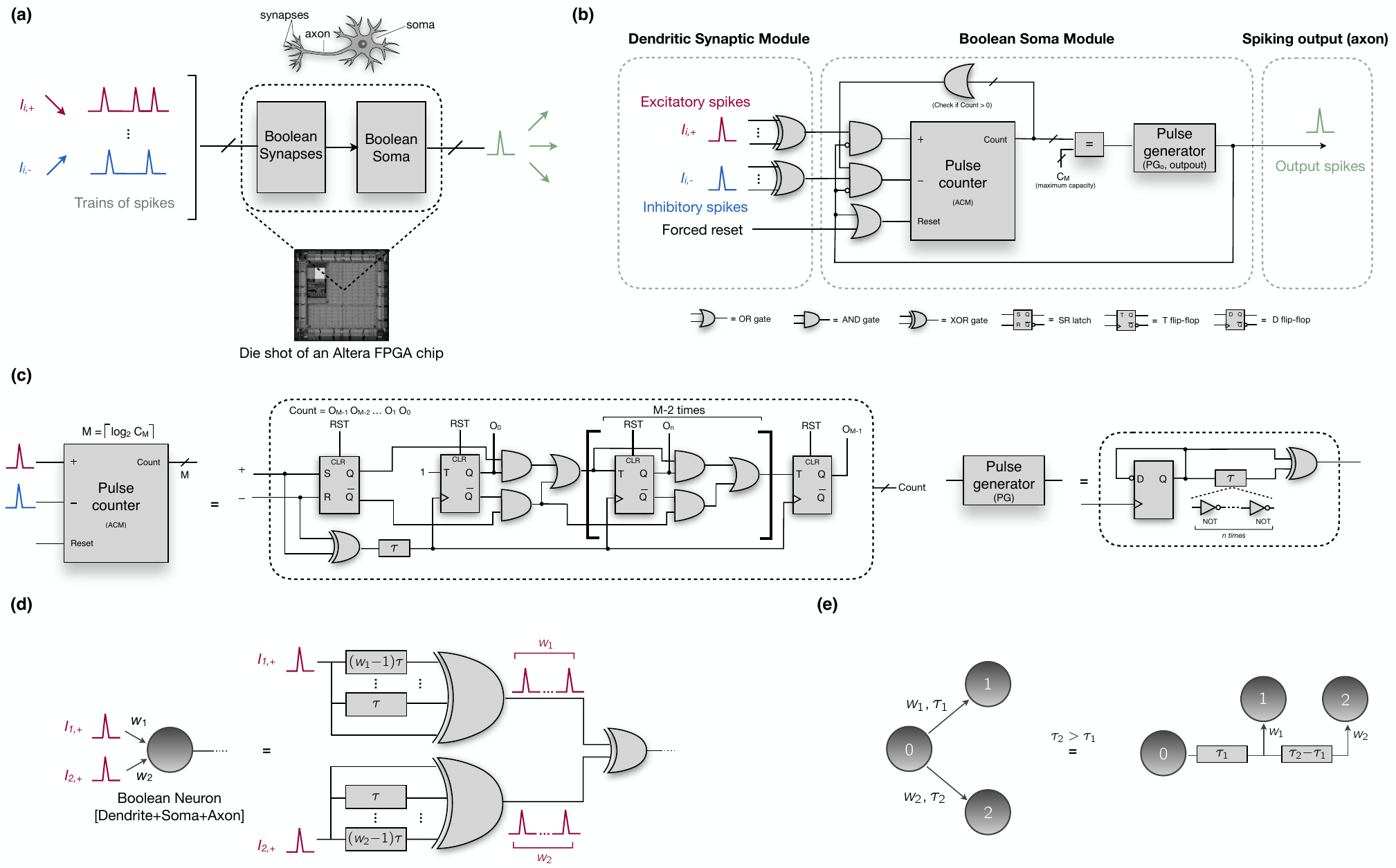}
\caption{Overview of the spiking Boolean neuron architecture. \textbf{(a)} Internal structure of the Boolean neurons depicting the interconnection between the two functional blocks implementing the soma and synapses \textbf{(b)} Detailed diagram of the Boolean neuron circuit. The soma module is composed of a bidirectional asynchronous counter module (ACM), modeling membrane depolarization and allowing for excitation and inhibition, and a pulse generator (PG$_\text{O}$) for producing fixed-width spikes. The counter capacity $C_M$, which is the threshold for triggering a postsynaptic signal, sets the number of T latches used by the counter. \textbf{(c)} ACM circuit architecture. A modified synchronous counter augmented with a mode-switching latch for bidirectional operation under asynchronous sparse spiking inputs. \textbf{(d)} Detailed weighted synaptic structure, illustrated on two excitatory synapses with weights $w_1$ and $w_2$. Each input signal is split into a number of branches equal to the weight magnitude, each with distinct propagation delays; these are then recombined using an XOR gate, yielding multiple output pulses. After weighting, signals from different sources are also aggregated using cascaded XOR gates. \textbf{(e)} Synaptic delay mechanism based on a single axon-inspired tapped delay line. Assuming delays $\tau_2$ and $\tau_1$, with $\tau_2 > \tau_1$, a delay of magnitude $\tau_2$ is implemented by extending the $\tau_1$ delay line with an incremental delay equal to their difference ($\tau_2 - \tau_1$).}
\label{fig:SBN}
\end{figure}

The main building block of our approach is an autonomous digital circuit that we dub a spiking Boolean neuron, shown in Fig{.}~\ref{fig:SBN}(a). Our neuron is organized around two main functional blocks : (i) the Boolean soma for accumulating a count analogous to a biological neuron's membrane charge, and implementing excitable behavior through comparison with a threshold, allowing for postsynaptic spikes resulting from presynaptic integration; (ii) the Boolean dendritic module responsible for combining spiking inputs from excitatory and inhibitory synapses, triggering increments or decrements to the counter. Synapses are formed by connecting a neuron's spiking output, analogous to an axon, to another neuron's dendritic module.

The Boolean soma is built around a pulse generator and a pulse counter, as illustrated in Fig{.}~\ref{fig:SBN}(b)-(c). The pulse generator is created with the design proposed in Ref{.}~\cite{Rosin2012}, having its principle of operation described in the Methods section. 
The pulse counter was implemented with a clockless (\emph{i.e.} asynchronous) bidirectional counting module (ACM) adapted into a custom-built spike-triggered circuit that we designed (see Methods). The inputs I$_{i,\pm}$ are for excitatory and inhibitatory presynaptic spikes, respectively. Excitatory spikes increment the counter, while inhibitory spikes decrement it. When the counter reaches its maximum value, the pulse generator is activated, triggering a new spike and resetting the count.  As a result, our proposed architecture builds an autonomous Boolean analogy of the integrate-and-fire (IF) behavior \cite{Gerstner_book_2014}.

Lastly, the dendritic circuit, shown in Fig{.}~\ref{fig:SBN}(d), consists of a wide exclusive-OR (XOR) gate, implementing a modulus-2 addition of logic inputs. This module combines spiking inputs coming from multiple Boolean neurons into individual increment or decrement signals to the ACM located in the soma. Hence, this module allows for an effective combination of input signals in continuous times, allowing for the construction of synapses through connections between spiking outputs and the dendritic XOR gates. 

The synaptic weights are directly implemented in the connection topology using a time-multiplexed strategy instead of being stored in digital registers. To implement an integer weight $w_i\in\mathbb{N}$ on an input spike, the signal is replicated across $w_i$ different delayed paths: each path adds a delay $k\tau$, with $k=0,\ldots,w_i-1$, and are recombined by an XOR gate. The resulting output signal has $w_i$ rising-edge transitions, which produce $\pm w_i$ increments of the ACM in the Boolean soma. Positive weights ($+w_i$) enhance the probability of firing a postsynaptic spike, \emph{i.e.} weighted excitatory coupling, while negative weights ($-w_i$) reduce this probability, implementing weighted inhibitory coupling (see Methods for details).

\subsubsection*{Autonomous Boolean Spiking Neural Network}
After defining a Boolean neuron clockless circuit, we aim to integrate these building blocks into a functional Boolean spiking neural network (B-SNN). In this work, the network was employed as a liquid state machine (LSM), a particular instance of the reservoir computing framework with a simplified training algorithm \cite{Maass2002}. The B-SNN is designed to promote rich spiking dynamics in response to input stimuli, seeking to maximize performance metrics for downstream machine learning tasks.

\begin{figure}[h]
    \centering
    \includegraphics[width=1\linewidth]{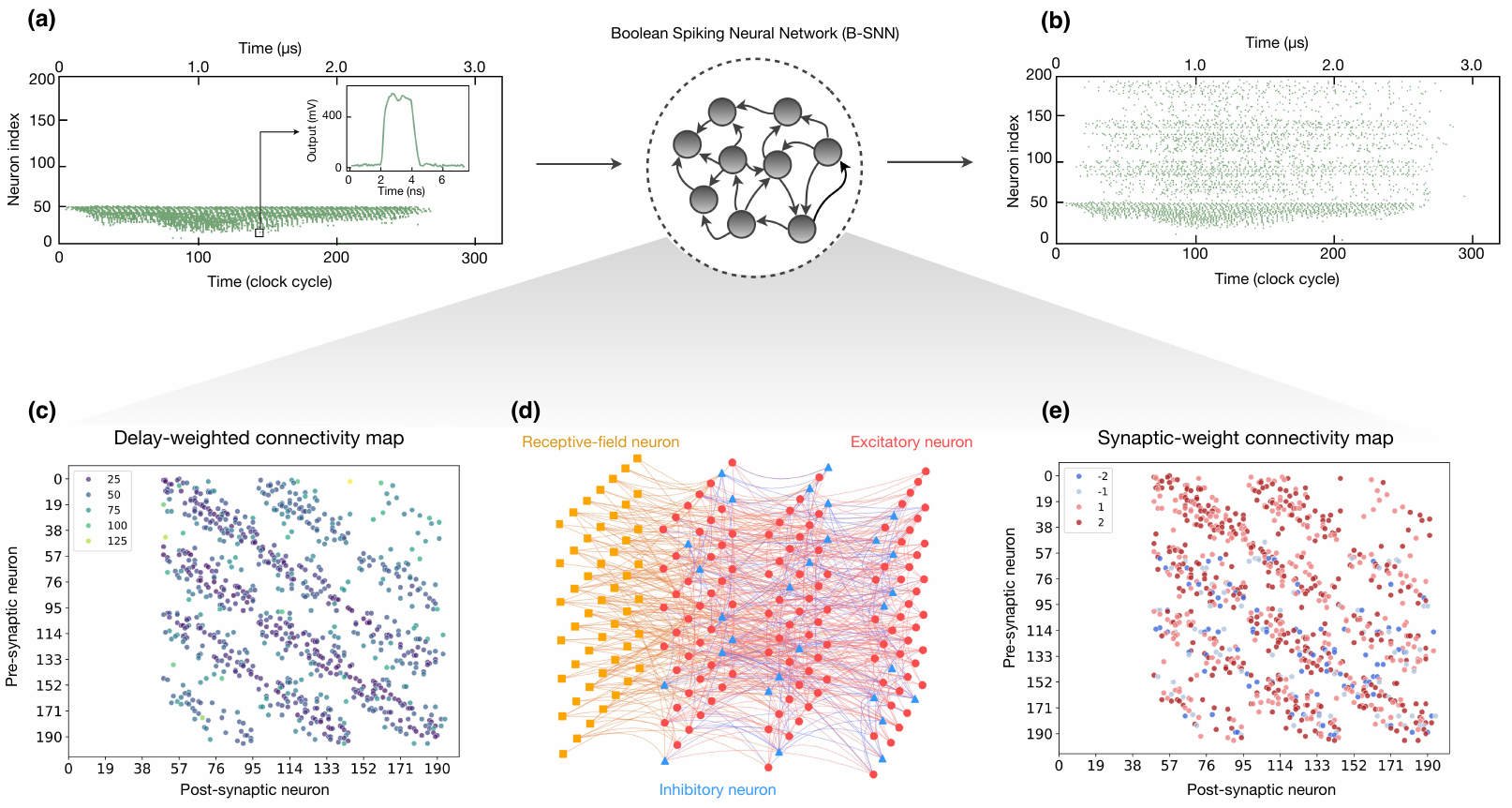}
    \caption{Overview of a typical B-SNN comprising 196 neurons arranged in a 7×7×4 grid, driven by spiking stimuli. \textbf{(a)} Raster plot over $3.2\,\mu\textrm{s}$ of a spiking input sample representing the spoken English digit ``one''. The highlighted region illustrates a representative spike with a full width at half maximum (FWHM) of $2.07\,\textrm{ns}$, measured using a high-speed oscilloscope. \textbf{(b)} Raster plot of the corresponding spiking output, observed over a $3.2\,\mu\textrm{s}$ time window. \textbf{(c)} Connection map illustrating the values in the network's synaptic delay-weighted adjacency matrix. \textbf{(d)} Network graph, with excitatory neurons depicted in red, inhibitory neurons in blue, and receptive neurons in yellow. $20\%$ of non-receptive neurons are inhibitory. \textbf{(e)} Connection map depicting the values in the network's synaptic strength-weighted adjacency matrix. Positive values correspond to excitatory synapses, whereas negative values correspond to inhibitory synapses.}
    \label{fig:topology_panel}
\end{figure}

In this study, the B-SNN provides a history-dependent nonlinear transformation, projecting low-dimensional spiking input sequences into a high-dimensional state space. Fig.~\ref{fig:topology_panel}(a) depicts a raster plot of a spike-encoded voice recording (spoken digit ``one''), consisting of 49 spike channels, used as a spiking stimulus to a sample B-SNN. Fig.~\ref{fig:topology_panel}(b) illustrates the corresponding spiking output raster plot, with a total of 196 spike channels, expanding the input dimensionality by a factor of 4. In this instance, the raster plots in Fig.~\ref{fig:topology_panel}(a)-(b) are discretized due to the limited time resolution of the measurement and input hardware, governed by the FPGA's clock. The reservoir itself operates in a clockless manner and generates spikes in continuous time, as evidenced by the highlighted measured pulse in Fig.~\ref{fig:topology_panel}(a), featuring a characteristic duration of \(2.07~\mathrm{ns}\), in contrast to the \(10\,\mathrm{ns}\) clock period used by the synchronous input and output devices. Further details on the acquisition and transmission of spiking data are provided in the Methods section.

Here, the B-SNN topology is realized by arranging neurons within a rectangular prism grid and interconnecting them according to a local connectivity rule. This architecture is inspired by the organization of neurons in cortical columns \cite{Maass2002} and enables reduced fan-in and fan-out, thereby lowering routing complexity and resource usage. The adopted grid dimensions are \(7 \times 7 \times 4\), resulting in a total of $196$ neurons. 
Given two neurons at positions \(a, b \in \mathbb{N}^3\) in the defined grid, the probability of a synaptic connection between them is defined as $P(a, b) = \Gamma \cdot \exp(\lVert a - b \rVert^2/\lambda^2)$, where $\Gamma$ and $\lambda$ are spatial connectivity parameters.

In accordance with Dale's principle \cite{strateDale}, a two-population model is adopted, where neurons are exclusively excitatory or inhibitory. That is, they emit only excitatory or only inhibitory postsynaptic spikes. This constraint is reflected in the choice of synaptic weights used in the network. In addition to excitatory and inhibitory neurons, a set of receptive neurons is defined, seeking to emulate receptive fields. We define receptive neurons as a subtype of excitatory neurons allowed to receive external inputs; they are confined to a restricted region of the B-SNN's grid. In this work, the receptive grid is defined as the first layer of the B-SNN's grid, acting as a secondary input layer. This approach is inspired by the work of Biswas et al. \cite{biswasTemporalSpatialReservoir2024a}, but differs from it by defining receptive neurons as a distinct subtype of excitatory neuron. The receptive layer, combined with the local connectivity rule, gives rise to receptive fields, a fundamental organizational principle of neurons in the auditory \cite{jenisonAuditorySpaceTimeReceptive2001} and visual \cite{ringachMappingReceptiveFields2004} cortex. Excitatory and inhibitory neurons are implemented with \(C_M=4\), establishing a four-spike threshold for postsynaptic firing. In contrast, receptive neurons are implemented with \(C_M=2\) for increased responsiveness to incoming stimuli.

The input layer is simplified due to the presence of the receptive layer. Each spike train present in the input data is mapped to a single receptive neuron, with a synaptic weight of 1 and no added propagation delay. Spike trains are assigned to the receptive layer based on spatial correspondence, aiming to extract key positional relations in the input. In this study, neighboring input channels correspond to adjacent frequency bands; accordingly, each channel is mapped to its corresponding index in the receptive grid, with the grid's indices flattened into a single linear index in row-column order.

With the inclusion of the receptive neurons, three neuron types are present in this model: (i) excitatory, (ii) inhibitory and (iii) receptive. Among the non-receptive neuron's, $20\%$ of neurons are defined as inhibitory.
Connectivity parameters were defined independently for each neuron-type pairing, where E, I, and R denote excitatory, inhibitory, and receptive neuron populations, respectively; thus, for example, E–I indicates connections from excitatory to inhibitory neurons. We set $\Gamma=0.3$ for E-I, I-E, R-E and R-I connections, and we define $\Gamma=0.15$ for E-E connections. Lastly, we set $\Gamma=0$ for all other connection types (I-I, E-R, I-R, R-R). The parameter \(\lambda\) is set to 2.2 for all pairings. The higher values of \(\Gamma\) for I-E and E-I connections, relative to E-E and I-I connections, play a crucial role in maintaining excitatory–inhibitory balance. A sample network defined over a 7×7×4 grid is illustrated in graph form in Fig.~\ref{fig:topology_panel}(d), highlighting the receptive layer and the local connectivity pattern.

Propagation delays in the B-SNN are implemented with chains of inverter pairs, each with a propagation delay $\tau_p = 560\pm 20$~ps. The synaptic delay between neurons in positions \(a\) and \(b\) is defined as \(D \cdot \lVert a - b \rVert\), with \(D = 20\,\tau_p\) and added Gaussian noise (\(\sigma = 3\,\tau_p\)) to increase variability. Fig.~\ref{fig:topology_panel}(c) represents a connectivity map, analogous to a connectome, of the synaptic delays obtained in a sample network. This magnitude of synaptic delay produces a clear temporal separation between spikes when measured with our custom-built time tagger (see Methods) offering a resolution of \(10\,\mathrm{ns}\), as the average synaptic delay for the shortest possible connection is equal to \(11.2\,\mathrm{ns}\); however, the parameter \(D\) can be reduced to speed-up the system dynamics. Synaptic weights between neurons are assigned values of $w_i = \{1,2\}$ with equal probability and implemented as described previously. Fig.~\ref{fig:topology_panel}(e) provides a connectivity map of the synaptic weights produced by this procedure. We note that by adopting this method to define synaptic weights and delays, these parameters are physically embedded in the network's structure rather than being stored in digital registers.

The selected level of network sparsity and inhibition reflect a compromise between providing sufficient inhibitory activity to stabilize neural dynamics and avoiding excessive suppression of network activity. As exemplified in Fig.~\ref{fig:topology_panel}(b), the obtained regime presents exploitable dynamics by maintaining a reasonable level of activity among non-receptive neurons, while still presenting a transient response to stimuli.

\subsection*{Integrated Machine Learning Pipeline with Boolean Spiking Dynamics}
To assess the potential of the developed Boolean spiking system for information processing, we chose the Spiking Heidelberg Digits (SHD) voice recognition dataset\cite{cramerHeidelbergSpikingData2022} as a benchmark. This task was selected because of its widespread use in the neuromorphic literature, providing a fair basis for comparison with other software and hardware implementations of SNNs. In this task, data samples are provided as spike-encoded sequences, requiring the B-SNN to effectively capture and process temporal dependencies in order to achieve high performance.

\begin{figure}[h!]
    \centering
    \includegraphics[width=1\linewidth]{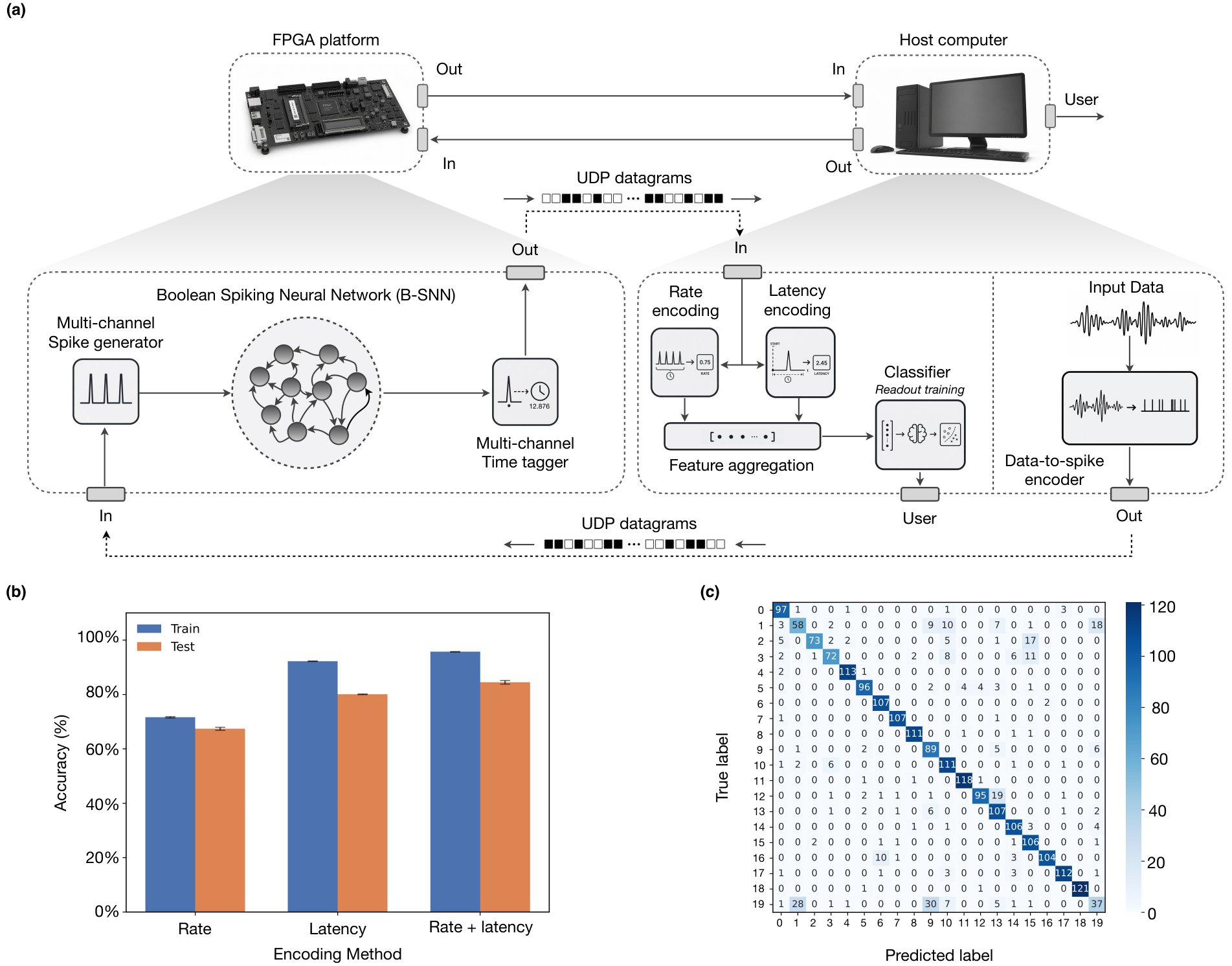}
    \caption{System pipeline and classification performance of a B-SNN for voice recognition. \textbf{(a)}~Overview of the complete experimental pipeline, consisting of event-driven input transmission to the FPGA, retrieval of Boolean-SNN spike events by a custom-built multi-channel time-tagger, and subsequent feature encoding for classification. \textbf{(b)}~Accuracies obtained using features derived from rate encoding, latency encoding, and their combination. Bars represent the mean over five runs; error bars indicate the standard deviation. The highest test accuracy ($84.50 \pm 0.67 \%$) was obtained using the combined encoding scheme. The B-SNN used has $196$ neurons. \textbf{(c)}~Confusion matrix on the SHD test set, comprising 2264 data points, for a single run using combined rate and latency encoding features. Labels 0–9 correspond to spoken digits in English, whereas labels 10–19 correspond to spoken digits in German.}
    \label{fig:shd_results_panel}
\end{figure}

Fig. \ref{fig:shd_results_panel}(a) illustrates the full pipeline from input data to predictions. Input data must provided as spike events occurring over a predefined number of channels. Event data is sent to a spike generator module in the FPGA through an Ethernet interface using the UDP protocol. This module transmits spike events to the Boolean-SNN, which yields a series of events for each observable Boolean neuron. Information about postsynaptic spike events in the Boolean-SNN is retrieved with a 200-channel time tagger with a $10\,\text{ns}$ time resolution included in the design. Retrieved data are transmitted to the host computer via UDP for post-processing, that is, encoding events into numerical embeddings, combining rate and latency encoding~\cite{guoNeuralCodingSpiking2021a} to produce feature vectors. Transmission and data retrieval are managed by a Python script running on a host computer, which is connected to the FPGA interface via a dedicated point-to-point Ethernet connection, with its transmission rate tuned to avoid packet loss.

Having built the feature vectors, a nonlinear readout layer is trained through multinomial logistic regression (softmax regression). We note that this model partitions the feature space with linear decision boundaries (hyperplanes)\cite{bishop2006pattern}. Model weights are optimized through the Limited-memory BFGS algorithm implemented in the \texttt{scikit-learn} \cite{scikit-learn} Python library. This model can then be used for inference by feeding new data points through the experimental pipeline.

\subsubsection*{Machine Learning Task Resolution}
The complete SHD training and test sets are fed to the experimental pipeline, producing feature vectors. Three experiments are performed for each data point, with the average of the produced feature vectors used for training and prediction. The logistic regression regularization parameter was tuned through a linear search, yielding the highest test accuracy with regularization parameter \(C=0.01\).

Table \ref{tab:shd_performance} presents a summary of the test accuracies achieved on the SHD dataset. The attained test accuracy ($84.50~\pm~0.67\%$) is comparable to that of the best performing analog computing method applied to SHD~\cite{dagostinoDenRAMNeuromorphicDendritic2024}($87.5\%$). Furthermore, our work represents, to the best of our knowledge, the first treatment of this problem with a physically implemented LSM. Nevertheless, a notable discrepancy remains between the test accuracy for the best performing software LSM ($89.3\%$) and the state-of-the-art software SNN ($96.26\%$), highlighting the trade-off introduced by the simplified training procedure, in which only the output (readout) layer is trained. An accuracy gap also persists between digital and analog approaches, stemming from challenges such as noise and network implementation constraints in analog architectures.

Our model has a total of 86.26k trainable parameters (196 neurons), significantly fewer than the state-of-the-art software LSM (256k parameters and 16k neurons) and analog accelerator (224k parameters). This presents an opportunity for bridging the current accuracy gap by expanding the network size, currently bounded by the 200-channel capacity of the time tagger.

A comparison of the accuracies obtained using rate encoding, latency encoding, and a combination of both, are presented in Fig.~\ref{fig:shd_results_panel}(b). A higher test accuracy is achieved when both the latency and rate features are used for classification, compared to using either of them in isolation, justifying the hybrid approach despite the increased feature vector size. However, latency encoding retains comparable accuracy, offering a viable alternative when model parsimony is prioritized. Analysis of the confusion matrix in Fig.~\ref{fig:shd_results_panel}(c) reveals that misclassifications occur more frequently between digits in different languages, which is consistent with the difficulty posed by phonetic similarities between English and German. A prominent example is the confusion between the English ``nine'' (label 9) and the German ``neun'' (label 19).

\begin{table}[ht]
\centering
\caption{Comparative classification accuracy and parameter count for the SHD task. All listed LSMs consist of fixed untrained recurrent SNNs, paired with trained readout (output) layers.}
\begin{tabular}{@{}lccc@{}}
\toprule
\textbf{Model} & \textbf{Test accuracy (\%)} & \textbf{Trainable parameters} & \textbf{Implementation type} \\
\midrule
Sun et al. 2025 \cite{sunParameterfreeAttentionalSpiking2025} & $96.26$ & $0.2\,\mathrm{M}$ & Software SNN \\
Schöne et al. 2024 \cite{schoneScalableEventbyeventProcessing2024} & $95.9$ & $0.4\,\mathrm{M}$ & Software SNN  \\
Baronig et al. 2024 \cite{baronigAdvancingSpatiotemporalProcessing2025} & $95.81 \pm 0.56$ & $0.45\,\mathrm{M}$ & Software state space model  \\
Hammouamri et al. 2023 \cite{hammouamriLearningDelaysSpiking2023} & $95.07 \pm 0.24$ & $0.2\,\mathrm{M}$ & Software SNN \\

\multicolumn{1}{c}{\vdots} & \vdots & \vdots & \vdots \\

Deckers et al. 2022 \cite{deckers_extended_2022} & $89.3$ & $256\,\mathrm{k}$ & Software LSM \\ % Calculated from paper info: 12 800 spike counts * 20 classes for logistic regression
Matinizadeh et al. 2025 \cite{matinizadeh_fully-configurable_2024} & $87.8$ & $-$ & FPGA-based SNN \\
D'Agostino et al. 2024 \cite{dagostinoDenRAMNeuromorphicDendritic2024} & $87.5$ & $224\,\mathrm{k}$ & Resistive RAM-based analog SNN \\
\textbf{Ours} & \textbf{84.50 $\pm$ 0.67} & \textbf{86.26\,k} & \textbf{FPGA-based quasi-analog LSM} \\
Cramer et al. 2022 \cite{cramerHeidelbergSpikingData2022} & $83.2 \pm 1.3$ & $-$ & Software SNN \\
Biswas et al. 2024 \cite{biswasTemporalSpatialReservoir2024a} & $77.8$ & $30\,\mathrm{k}$ & Software LSM \\
Carpegna et al. 2025 \cite{carpegna_spiker_2025} & $72.99$ & $-$ & FPGA-based SNN \\
\midrule
\multicolumn{3}{@{}l}{Reference non-spiking architectures (Cramer et al., 2022 \cite{cramerHeidelbergSpikingData2022})} \\
\midrule
CNN & $92.4 \pm 0.7$ & $-$ & Software \\
LSTM & $89.0 \pm 0.2$ & $-$ & Software \\
Linear SVM & $56.0 \pm 0.4$ & $-$ & Software \\
\bottomrule
\end{tabular}

\vspace{0.5em}
\label{tab:shd_performance}
\end{table}

\section*{Discussion}
The developed Boolean neuron architecture proves capable of reproducing charge accumulation and excitable behavior, replicating the fundamental dynamics of IF neurons. It also shows the capability of handling both excitatory and inhibitory inputs, allowing for the creation of networks with balanced activity, making their use viable as LSMs. Furthermore, the system operates on a fast timescale. This is evidenced by the measured spike duration of approximately \(2.07\,\mathrm{ns}\), driving an inherently parallel evolution of the system over similar temporal dynamics. In contrast, the used FPGA operates at a default clock frequency of \(50\,\mathrm{MHz}\), corresponding to a \(20\,\mathrm{ns}\) clock period, which is an order of magnitude slower than the observed spike duration.

However, the simplification of neuron dynamics in the creation of the Boolean neuron leads to limitations. Namely, the discretization of propagation delays and synaptic weights, and the lack of charge decay in the absence of inhibitory inputs. Additionally, the combination of input spikes may present non-ideal behavior, with the possibility of spikes being miscounted due to infavorable timings, specially when input spikes closely overlap. Synaptic weights and delays are currently fixed during synthesis, which precludes plasticity mechanisms unless the architecture is modified, for example through multiplexing between weighting structures and delay lines. Exploring plasticity and full-network training for the developed architecture remains an important avenue for future investigation.

Despite these limitations, the module is experimentally shown to exhibit exploitable dynamics. In particular, within the LSM framework, it is demonstrated that the SHD task can be solved with remarkable accuracy.
A network implemented on a low-cost FPGA achieved a test accuracy of $84.50 \pm 0.67 \%$ on the test dataset, showing only a small gap to the current state-of-the-art analog computing method \cite{dagostinoDenRAMNeuromorphicDendritic2024} for the SHD dataset. Additionally, it also constitutes the first treatment of the SHD dataset with a physically implemented LSM.

Two main design choices were made with the intent of achieving this level of performance. Firstly, a receptive layer was integrated into the Boolean SNN, aiming to exploit spatial structure in the data by triggering similar dynamical responses for closely related input channels. Secondly, an output encoding method was devised to improve performance by combining both rate and latency encoding schemes, thereby integrating complementary information to enhance generalization. In particular, the latency encoding leverages the timing of multiple spikes, rather than relying solely on the first spike, as in time-to-first-spike (TTFS) encoding~\cite{guoNeuralCodingSpiking2021a}. This enables high-performing classifiers with a small number of liquid neurons, as demonstrated here. This benefit comes at the cost of requiring the predefinition of observation windows tailored to the task being solved, as well as the use of a number of features higher than the number of liquid neurons. Nonetheless, this represents a reasonable trade-off given the achieved performance, provided that the number of parameters in the readout layer does not constitute a significant constraint. Thus, in addition to demonstrating the utility of the novel Boolean neuron module, the attained results highlight possibilities for enhancing LSM architectures through modifications to the network topology and adoption of alternative output encoding schemes. Such exploration may help to reduce the current performance gap between physical LSM implementations and state-of-the-art machine learning approaches, albeit at the cost of increased complexity and manual tuning.

To evaluate the scalability of our architecture, we analyze the number of logic elements required after synthesis as we varied the number of grid layers in the B-SNN topology. Grid dimensions ranged from 7$\times$7$\times$2 to 7$\times$7$\times$5. The smallest configuration, containing $98$ neurons, required $12{,}329$ logic elements, whereas the largest implemented network, containing $245$ neurons, required $45{,}871$ logic elements. Supporting infrastructure — including experiment control, the time tagger, spike generator, and Ethernet interface — introduced an approximately constant overhead of $8{,}644$ logic elements and $1.14$ Mib of memory.
For a B-SNN containing $N_n$ neurons (with medium size), we uncover an empirical scaling law $N_{LE}=15.40 N_n^{1.46}$ (determination coefficient for the regressed scaling model is $R^2>0.995$) for the required number of logic elements. Extrapolating this trend suggests that the FPGA used in this work with $114{,}480$ logic elements could support networks containing up to $447$ neurons. In comparison, a high-end FPGA chip with approximately $10.2$ million logic elements could support networks with up to $9{,}696$ neurons with our proposed design.

Despite these encouraging scaling estimates, large-scale implementations still face several practical challenges. In particular, the FPGA design software suite was unable to place properly and route networks larger than 7$\times$7$\times$5 ($245$ Boolean neurons). To overcome this limitation, one possible strategy would be to develop custom placement-and-routing methods for the proposed architecture instead of relying on the general-purpose algorithms provided by the conventional FPGA toolchains. Another limitation is related to the B-SNN observability: our present implementation of the time tagger used to collect spiking events supports up to $200$ parallel measurement channels. This limitation stems from the data-buffer IPs used in the implementation. Therefore, modifications to the current time tagger or alternative acquisition strategies, \emph{e.g.} time multiplexing, would be necessary to monitor larger networks. Another straightforward option would be to build several time taggers working synchronously on probing different parts of the B-SNN.

An additional important consideration is the achievable bandwidth of a Boolean spiking neural network. In our design, feature vectors are generated from autonomous spiking events using time windows of $10.24$ $\mu$s (\textit{i.e.} $1024$ clock cycles), yielding a theoretical maximum bandwidth of $97.66$ kHz for machine learning applications. Transitioning to higher-performance FPGA devices with faster timing resolution would enable higher sampling rates of autonomous spiking activity to extract features, hence further compressing the overall system timescale. It is worth noting that faster network dynamics can also be achieved by reducing the delay-line lengths, which have been purposefully chosen to slow-down the B-SNN dynamics to accommodate slow sampling rates in our experiments. Another bandwidth limitation is directly linked to the digital components used in the presented prototype, namely the host computer and Ethernet interface. This is a major processing bottleneck, because to fully exploit the capacity of a B-SNN, all modules would have to operate on comparable (nanosecond) timescales, which is a challenging requirement for software. Therefore, the integration of a high-performance physical readout layer is essential to fully leverage the high-speed processing capabilities of our architecture on FPGA.

In addition to task performance, power consumption represents a central concern in neuromorphic computing. Accordingly, we develop an analysis centered on the implemented LSM, that is, the proposed SNN architecture. Gate-level simulations were performed to estimate the energy consumption of the LSM, excluding contributions from input and output devices, resulting in a total thermal power of $192.37$~mW. The average energy cost amounts to $1.99$~$\mu$J given that $20$ input samples require $206.79$~$\mu$s to be processed in the performed simulation. 

The obtained estimated total power consumption and energy consumption per SHD recording are remarkable for an FPGA implementation. Among digital SNN hardware applied to the SHD dataset, Spiker+ attains a power consumption of 430 mW, consuming 230 $\mu$J per audio recording~\cite{carpegna_spiker_2025}, while 1.629 W peak dynamic power is reported for QUANTISENC~\cite{matinizadeh_fully-configurable_2024}. Two main factors likely contribute to the attained low energy consumption: the low switching rates triggered by sparse spiking activity, minimizing dynamic power consumption, and the low latency (10.24 $\mu$s observation windows) enabled by high-speed dynamics.

This result attests the advantage of exploiting clockless continuous-time systems in place of a conventional digital design, albeit at the expense of the determinism characteristic of numerical methods. However, it must be remarked that the level of efficiency remains limited by the used substrate. The resistive RAM-based proposal by D'Agostino et al. achieves a power consumption of 8.41 $\mu$W~\cite{dagostinoDenRAMNeuromorphicDendritic2024}, highlighting a substantial gap with respect to FPGA-based implementations.

Nonetheless, we note that these systems operate at different timescales. The resistive-RAM architecture~\cite{dagostinoDenRAMNeuromorphicDendritic2024} is engineered for real-time operation on a biological timescale, whereas our approach prioritizes high throughput by operating at a timescale compressed by a factor of $200{,}000$. Consequently, the two approaches are optimized for different application regimes.

On the whole, our work presents a promising direction for exploration mainly due to its applicability to conventional reconfigurable devices and, more broadly, CMOS integrated circuits. Moreover, power estimates suggest that by exploiting quasi-analog dynamics, this work achieves a promising route toward energy-efficient computation with digital circuits. Hence, our work offers a widely accessible platform for implementing low-power continuous-time spiking dynamics.

\section*{Conclusion}
We have shown that asynchronous Boolean networks can be engineered to exhibit key neuronal properties, such as excitatory and inhibitory interactions and an integrate–and–fire–like mechanism, while supporting configurable synaptic weights and delays. It enables the realization of large-scale networks of Boolean spiking neurons useful for general-purpose neuromorphic computing. Unlike conventional digital approaches, our proposed clockless architecture exploits quasi-analog, continuous-time dynamics arising from the intrinsic response of the chip transistors, and allows event-driven computation directly embedded in the network structure. This results in massively parallel operation at nanosecond timescales.

When used in a machine-learning pipeline, within the liquid state machine (LSM) paradigm, our system achieves performance comparable to state-of-the-art analog neuromorphic hardware while remaining implementable on commercially available Field-Programmable Gate Array (FPGA) platforms. It makes for a compact, low-cost experimental realization that uniquely combines performance, speed, and broad accessibility. Power estimates further indicate favorable energy efficiency, demonstrating a two-order-of-magnitude improvement over digital FPGA implementations of spiking neural networks. However, integrating the readout layer on hardware remains an important direction for full system-level realization.

In all, we show that clockless (asynchronous) digital circuits are a practical and scalable platform for neuromorphic computing, bridging the gap between conventional digital processors and dedicated analog accelerators. They open the door to repurposing existing reconfigurable hardware for energy-efficient neuromorphic systems.

\section*{Methods}
\subsection*{Autonomous Boolean Design}
To implement the proposed autonomous Boolean network architecture, primitive blocks were designed, aiming to achieve the desired excitable behavior. Two main components are used to construct the Boolean soma: a pulse generator, with the design described in Ref{.}~\cite{Rosin2012}, and a custom spike counter. In addition to the soma, the neuron contains a Boolean dendritic module, responsible for combining inputs from different sources through physical connections.

The pulse generator uses an clockless D flip-flop triggered by an external rising Boolean transition (rising edge) on its clock input. Its output $\overline{Q}$ is fed back to its $D$ input to ensure the generation of a Boolean transition at the output $Q$ whenever a rising edge is detected at the clock input. The output $Q$ is then fed to two delay lines of different duration made of linear chains of an even number of clockless inverters (NOT gates), combining the propagation delay of FPGA logic elements to construct modules with an arbitrary propagation delay. The outputs of each line are combined with a Boolean XOR gate, producing an output pulse with length equal to the duration difference between the delay lines. Thus, the shortest pulse that can be generated has a length equal to the propagation delay through a pair of NOT inverters (typically $\tau_p = 560\pm 20$ ps in an EP4CE115F29C7 FPGA chip). The pulse generator (PG$_\text{S}$) will produce the Boolean neuron spike, which is also used to prevent counting while a spike is being fired. Here, a dedicated pulse generator is not used to implement the refractory period as described in Ref{.}~\cite{Rosin2012}. Nevertheless, the neuronal dynamics still exhibit a minimum inter-spike interval, imposed by the minimum number of sufficiently long pulses required to increment the counter to its threshold. Pulses shorter than a minimum width may be filtered by the logic gates used, failing to reliably trigger the counter; empirically, a pulse width of $2.07$ ns is sufficient to ensure consistent counting. An absolute refractory period could optionally be implemented by adding a separate pulse counter to the circuit, in order to block inputs during a period longer than the output pulse duration.

The spike counter circuit is derived from a conventional synchronous digital counter and adapted for asynchronous operation triggered by short-duration pulses. The capacity $C_M$ of the counter is our proposed Boolean translation of the membrane polarisation effect at the somatic level in biological neurons\cite{Gerstner_book_2014}: The larger the capacity $C_M$, the more excitatory spikes are necessary to trigger a post-synaptic pulse. The inhibitory spikes, to the contrary, will prevent the asynchronous counting module from reaching it maximum capacity and will tend to drive it to its lowest level at $C_M = 0$. Given these inputs, the increment / decrement mode is defined by an SR flip-flop, toggled by excitatory and inhibitory spikes. To prevent the counter from being triggered in the incorrect mode, excitatory and inhibitory spikes are combined in a mutually exclusive manner using an XOR gate, and an additional delay of \(2 \tau_p\) is introduced via delay line. It is important to note that a wide OR gate is used to detect when the count is nonzero, which is a necessary condition for allowing the counter to be decremented, thereby preventing an underflow error.

Besides the blocks constituting the Boolean soma, we propose a Boolean dendritic module to combine spikes coming from different sources. In an FPGA, this block is implemented with a wide XOR gate, consisting of a cascade of gates if the in-degree (fan-in) exceeds the capacity of a single logic element.
The choice for XOR gates is motivated by the continuous time dynamics of the input signals, since, with an XOR gate, two partially overlapping pulses produce two distinct pulses. In contrast, an OR gate would merge any two overlapping pulses into a single pulse, thereby inducing miscounting through a single increment or decrement.

A similar design is employed to implement synaptic weights. To realize weighted connections, a spike channel is divided into \(N_b\) branches, each with a distinct propagation delay, and the resulting signals are recombined using an XOR gate. This procedure produces \(N_b\) rising-edge transitions, corresponding to an increment or decrement of \(N_b\). Experimentally, this approach yielded reliable counting when the propagation delay difference between branches is set to \(5\tau_p\), which was used in the final network.

The operation of the proposed autonomous Boolean design relies on the sparsity of spike events expected in its intended use cases, as closely spaced spikes may lead to ill-defined behavior, an inherent limitation of the proposed metastable architecture. Nevertheless, such rare anomalies can be regarded as noise due to their low occurrence in sparsely active networks. The limited impact of these events is supported by the low variance reported in the experimental results. Moreover, the spike duration must be sufficiently long to avoid elimination by the implemented components. Hence, we set the ideal spike duration to \(4\,\tau_p = 2.24\,\mathrm{ns}\), which is sufficient for reliable spike counting and prevents suppression by the synaptic structures.

\subsection*{Liquid State Machine with Boolean Spiking Dynamics}
Custom digital hardware, hereafter referred to as the spike generator, is employed to produce input spikes. As this circuits is implemented with synchronous digital logic, spikes are produced at discrete time steps. In particular, we adopt a time step of $10$ ns, used both for driving inputs and sampling outputs. To achieve this in the FPGA, the default 50 MHz clock frequency is doubled to 100 MHz through a PLL. Nonetheless, it is important to remark that reservoir dynamics remain quasi-analog, with the clock signal only being used to drive auxiliary circuits.
Besides the signal source, a physical input layer provides an interface between the spike generator and the reservoir, connecting each channel of the generator to a subset of network neurons. In the studied architecture, this subset corresponds to the receptive neurons.

As soon as an input sequence starts being generated by the spike generator, postsynaptic spike responses of neurons in the reservoir are monitored. This is achieved using a custom time tagger, which detects the presence of spikes within a sequence of non-overlapping time intervals of fixed duration. Specifically, our time tagger segments time into consecutive 10 ns intervals, collectively forming an observation window. Within each interval, the presence or absence of a spike is recorded, yielding a discrete representation of neural activity. This result can be represented as a binary observation matrix \(O \in \{0,1\}^{T_{out} \times N}\), where \(N\) denotes the number of observed neurons and \(T_{out}\) the size of the observation window.
The time counter used by the time tagger is restarted at the start of every observation window, which is necessary for the latency encoding method to yield consistent results.

Hence, for each data point processed, first the spike-encoded input is transmitted to a buffer in the FPGA. Then, the input stored in the FPGA's internal memory is read by the spike generator, which produces spike trains mapped by a physical input layer to the reservoir. When the spike generator is started, the time tagger is activated simulataneously to monitor generated spike events. Each observation is transmitted to the host computer via an Ethernet connection using UDP packets.

Subsequently, the observations are transformed into real-valued vectors to serve as input to a classifier. In this work, we adopt both rate-based and delay-based encoding techniques to construct these vectors, integrating information encoded in both firing rates and precise spike timing.
Two distinct feature sets are derived from the spike rates: the absolute number of spikes for each neuron, and the relative number of spikes, obtained by dividing each neuron's spike count by the total number of spikes in the observation window. Scaling is also applied by dividing the feature means by the standard deviation, which aims to ease convergence during training.
Precise spike timing is incorporated into the model by including the time of occurrence of each neuron's first 20 spikes into the feature vector. If no spike is observed for a certain neuron, the corresponding null value is replaced by the 0.99 quantile of spike times within the observation window. Each feature is scaled by subtracting its median and dividing by its interquartile range, providing a robust normalization that mitigates the influence of extreme values arising from imputed absent spike events.

By combining both the rate and temporal encoding, a single feature vector \(\mathbf{x_i} \in \mathbb{R}^{22 N}\) is obtained for each observation window \(O_i \in \{0,1\}^{T \times N}\) of duration \(T\). Each observation window, and therefore each feature vector, corresponds to a data sample to be classified, and is thus associated with a class label \(y_i \in \mathbb{N}\). Consequently, given a dataset consisting of \(m\) samples, the target vector is \(y = (y_1, \dots, y_m)^\top \in \mathbb{R}^m\) and the design matrix is \(\mathbf{X} = [\mathbf{x_1}^\top; \dots; \mathbf{x_m}^\top] \in \mathbb{R}^{m \times (22 N)}\).

A classifier with linear decision boundaries is employed as the readout layer, with the extraction of temporal patterns and the transformation into a high-dimensional nonlinear representation delegated to the reservoir. In this instance, since multiclass classification is performed, multinomial logistic regression is applied.

This model is described in Eq.~(\ref{eq:softmax_classif}), $\mathbf{x} \in \mathbb{R}^d$ denotes the feature vector, where $d$ is the number of features. 
$y \in \{1, \dots, K\}$ represents the class label, and $k$ indexes a particular class among the $K$ possible classes. 
For each class $k$, a weight vector $\mathbf{w_k} \in \mathbb{R}^d$ and bias term $b_k \in \mathbb{R}$ are introduced; together they constitute the model's trainable parameters. The softmax function normalizes the scores across all classes to produce the conditional probability $P(y = k \mid \textbf{x})$. The final class label is assigned to the category $k$ with the highest conditional probability.

\begin{equation}
P(y = k \mid \mathbf{x}) =
\frac{\exp(\mathbf{w}_k^\top \mathbf{x} + b_k)}
     {\sum_{j=1}^{K} \exp(\mathbf{w}_j^\top \mathbf{x} + b_j)} \label{eq:softmax_classif}
\end{equation}

The parameters \(\mathbf{W} = (\mathbf{w_0}, \dots, \mathbf{w_K})\) and \(\mathbf{b} = (b_0, \dots, b_K)^\top\) are tuned by minimizing a cost function. For multinomial logistic regression, we adopt the cross-entropy loss, with an added regularization parameter $C$. Given a dataset with $m$ samples, the cost function is detailed in Eq.~(\ref{eq:softmax_loss}).
\begin{equation}
J(\mathbf{W}, \mathbf{b})
= -\frac{1}{m} \sum_{i=1}^m \log \big( P(y_i \mid \mathbf{x}_i; \mathbf{W}, \mathbf{b}) \big)
\;+\;
\frac{1}{2C} (\lVert \mathbf{W} \rVert^{2} + \lVert \mathbf{b} \rVert^{2}) \label{eq:softmax_loss}
\end{equation}

For parameter training, the Limited-memory Broyden-Fletcher-Goldfarb-Shanno (L-BFGS) algorithm \cite{liu_limited_1989} is chosen over standard gradient-based approaches due to its faster convergence and widespread adoption in machine learning libraries. In particular, we rely on the \texttt{scikit-learn} library \cite{scikit-learn} implementation. As a second-order method, L-BFGS approximates Newton's method for optimization by computing an approximate inverse Hessian, enabling fast convergence compared to first-order methods such as gradient descent.

\subsection*{Experiments and Implementation with Field-Programmable-Gate-Arrays}
All experiments were conducted on an Altera DE2-115 development board, containing a Cyclone® IV EP4CE115F29C7 FPGA. This device provides a total of $114{,}480$ logic elements. Each logic element (LE) contains a look-up table with a fan-in of four, enabling the implementation of combinational logic with up to four inputs, as well as a D flip-flop for storage elements. Intel Quartus Prime Version 25.1 was used to synthesize the hardware descriptions, written in Verilog, into the board. Synthesis directives are employed to preserve the integrity of the asynchronous design during optimization, preventing unintended modifications by the synthesis tool.

In order to monitor the network, a time tagger is synthesized into the FPGA, allowing up to $200$ measuring channels (or probes) to be internally connected to neuron outputs, with a time resolution of $10\,\textrm{ns}$. This design was inspired by the 4-channel FPGA time tagger proposed by Gamari et al.~\cite{gamari_inexpensive_2014}, available under the GNU General Public License v3.0. Departing from this base architecture, we establish a significant expansion of the number of channels from $4$ to $200$. As for input generation, the synthesized system also incorporates a custom-built spike generator, supporting up to $128$ spiking input channels.
Additionally, we use a Tektronix MSO64B oscilloscope with $6$ GHz analog bandwidth and sampling rate of $50$ GS/s to directly observe analog spikes generated by B-SNN.

It was observed that 22 logic elements are needed to implement an isolated Boolean neuron with a 4-bit ACM. However, since the delay lines between neurons are composed by chained NOT gates, they also require substantial resources, which scale with delay magnitude. Furthermore, synaptic module resource consumption scales with the presynaptic neuron count and synaptic weight magnitudes. Hence, the reservoir size is largely determined by the chosen network topology. In contrast, the auxiliary infrastructure requires a constant amount of logic elements and memory blocks, subject to small variations due to optimization during synthesis. The Ethernet interface requires 4730 logic elements and 112 M9K memory blocks (0.866 Mib). The time tagger requires 2862 logic elements and 29 M9K blocks (0.25 Mib), while the spike generator requires 676 logic elements and 4 M9K blocks (24.6 Kib). Components for integrating and controlling these modules require 376 additional logic elements.

\subsection*{Numerical Simulations}
In order to evaluate the ideal behavior of the developed circuit, simulations were performed with Questa*-Intel FPGA Edition 2025.2, a widely utilized simulation tool for hardware description languages, enabling functional and timing analysis of digital circuits, as well as gate-level simulations. In this context, this tool is used to export waveforms obtained after stimulating sets of artificial neurons with different inputs.

This procedure captures the main properties of the neuron circuit and the reservoirs built by associating them, while handling only digital signals. This is viable since only the time-delay properties and Boolean logic are needed to produce the desired spiking signals, allowing for a good approximation without an analog model.

Nevertheless, propagation delay specification for the circuit's logic gates in the Verilog hardware description is crucial for obtaining the expected dynamics in functional simulations. In this context, the propagation delay is assumed to be constant at 280 ps, reflecting the average propagation delay measured for NOT gates in the FPGA under consideration.

Gate-level simulations were performed to obtain waveforms with precise timing, using descriptions of the synthesized circuit generated in Intel Quartus Prime for the target FPGA model. These simulations served as input to the Power Analyzer tool in Quartus, yielding estimates of thermal power dissipation. To compute metrics on the synthesized circuit’s dynamics, 20 randomly sampled input instances from the SHD dataset were applied to the simulated reservoir, and the resulting activity was used to estimate power consumption.

This method allows for simulating and performing experiments on the same hardware descriptions. However, the tools used for interfacing with the reservoir are not included in the simulations, as inputs and outputs can be handled directly by the simulation testbench.

\subsection*{Benchmark task and preprocessing}
To evaluate the applicability of the proposed architecture to practical tasks, the Spiking Heidelberg Digits (SHD) dataset was chosen. The SHD dataset \cite{cramerHeidelbergSpikingData2022} consists of $10{,}420$ audio recordings, each containing a spoken digit from 0 to 9 in English or German. Labels 0–9 correspond to the spoken digits in English, while labels 10–19 correspond to the same digits (0–9) in German. A subset of $2{,}264$ files are set apart for testing, while the remaining $8{,}156$ are used for training.
SHD provides data in a spike-encoded format, enabling the direct use of the provided spike events in neuromorphic architectures, reducing the effect of input preprocessing on the final test accuracy. Nevertheless, we adopt a reduction in the number of input channels and time discretization through binning, for compatibility with the custom spike generator integrated into the FPGA platform.

The original set of spike-encoded audio files contains 700 input channels in continuous time. First, time must be discretized due to the digital input procedure; therefore, spike are grouped into 2 ms bins, yielding a two-dimensional Boolean array representation. To reduce input dimensionality, adjacent channels are combined into 49 channels via Boolean addition. This input compression is known to improve generalization \cite{cramerHeidelbergSpikingData2022}. Lastly, the training set is augmented by introducing a Gaussian channel jitter with a standard deviation of 20 to all samples prior to channel merging. The augmented samples are appended to the original dataset, doubling the size of the training set.

We report the average test accuracy and standard deviation obtained over five experimental runs using the same reservoir instance. The inherently stochastic behavior of the network leads to the observed variability in results, in contrast to the usual reported variance of software-based reservoirs, which stems from repeating experiments with different reservoir instances.

\bibliography{Biblio}

\section*{Acknowledgements}

The authors would like to gratefully acknowledge the financial support of the Conseil of Region Grand-Est, the AirForce Office for Scientific Research (AFOSR) and Office for Naval Research (ONR) through grant FA8655-22-1-7031, the Agence Nationale de la Recherche (ANR) through grant ANR-24-CE24-4767 (AATLAS), and the Intel FPGA Academic Program. 

\section*{Author contributions statement}

D.R. designed the study, conceived the experiment and managed the overall project. E.O.G was supervised by D.R for this work. D.R. and E.O.G. co-designed the Boolean spiking neural network architecture. E.O.G performed the numerical simulations in the Questa environment and the experiments with the FPGA platform. All the authors discussed the results and analysed the data. D.R. and E.O.G. wrote the core text of the manuscript and E.O.G the methods section. All the authors reviewed and contributed to the improvement of the manuscript. 

\section*{Competing interests}
The authors declare no conflicts of interest.

\end{document}